%% file: iclr2027_conference.tex
\documentclass{article} 
\usepackage{iclr2027_conference,times}

\input{math_commands.tex}

\usepackage{hyperref}
\usepackage{url}

\usepackage{multirow}
\usepackage{makecell}
\usepackage{inconsolata}
\usepackage[english]{babel}
\usepackage{xcolor} 
\usepackage{enumerate}
\usepackage{enumitem}
\usepackage{pifont}     
\usepackage{array}
\usepackage{makecell}   
\usepackage{hyperref}
\usepackage{xcolor}
\usepackage{tcolorbox}
\usepackage{caption}
\usepackage{lipsum}

\usepackage{microtype}
\usepackage{graphicx}
\usepackage{subcaption}
\usepackage{booktabs}

\usepackage{amsmath}
\usepackage{amssymb}
\usepackage{mathtools}
\usepackage{amsthm}

\usepackage[T1]{fontenc}        

\definecolor{mainblue}{rgb}{0.0, 0.0, 1.0}
\definecolor{mainred}{rgb}{1.0, 0.0, 0.0}
\definecolor{textgray}{gray}{0.2} 

\newcommand{\cmark}{\ding{51}}
\newcommand{\xmark}{\ding{55}}
\newcommand{\fstar}{\ding{72}} 

\newcommand{\starl}{\fstar}
\newcommand{\starm}{\fstar\fstar}
\newcommand{\starh}{\fstar\fstar\fstar}

\title{LSR-Ben: A Logical and Scientific Reasoning Benchmark for Evaluating Process Reward Models}

\author{Zhouhao Sun\textsuperscript{1}, Xuan Zhang\textsuperscript{1}, Xiao Ding\textsuperscript{1$\ast$}, Bibo Cai\textsuperscript{1}, Li Du\textsuperscript{2}, Kai Xiong\textsuperscript{1}, Xinran Dai\textsuperscript{1}, 
\\
\textbf{Fei Zhang\textsuperscript{1}, Weidi Tang\textsuperscript{1}, Zhiyuan Kan\textsuperscript{1}, Yang Zhao\textsuperscript{1}, Bing Qin\textsuperscript{1}, Ting Liu\textsuperscript{1}}
\\
${}^{1}$Harbin Institude of Technology, ${}^{2}$Beijing Academy of Artificial Intelligence}

\newcommand\blfootnote[1]{%
  \begingroup
  \renewcommand\thefootnote{}\footnotetext{#1}%
  \addtocounter{footnote}{-1}%
  \endgroup
}

\iclrfinalcopy 

\begin{document}
\maketitle

\begin{abstract}
Currently, process reward models (PRMs) have exhibited remarkable potential for test-time scaling. 
Since large language models (LLMs) regularly generate flawed intermediate reasoning steps when tackling a broad spectrum of reasoning and decision-making tasks, PRMs are required to possess capabilities for detecting process-level errors in real-world scenarios. 
However, existing benchmarks primarily focus on mathematical reasoning, thereby failing to comprehensively evaluate the error detection ability of PRMs across diverse reasoning scenarios. 
To mitigate this gap, we introduce LSR-Ben, a process-level benchmark specifically designed for assessing PRM's performance across two primary reasoning domains (scientific and logical reasoning) and nine subdomains. 
We conduct extensive experiments on a diverse set of 22 models, encompassing both PRMs and LLMs, and derive two key findings: (1) In domains beyond mathematical reasoning, the error-detection ability of existing PRMs and LLMs is found to be markedly weaker by comparison.
(2) In general,  LLMs exhibit a tendency toward over-identification of errors compared to PRMs, whereas PRMs exhibit an inherent tendency to overlook errors compared to LLMs. 
We hope LSR-Ben can foster future researches on PRMs for broader domains, thereby enhancing the reasoning capabilities of LLMs. 
\blfootnote{$^\ast$ Corresponding Author.}
\end{abstract}

\section{Introduction}

With the development of test-time scaling, large language models (LLMs) such as Gemini-3 \cite{comanici2025gemini} and GPT-5 \cite{singh2025openai} have achieved remarkable progress in complex mathematical problem-solving. Against this backdrop, process reward models (PRMs) play a pivotal role in facilitating this paradigm \cite{chen2025better,zheng2025survey,chen2025discriminative}. Specifically, PRMs are capable of delivering fine-grained, step-wise evaluations, which in turn enables LLMs to explore a broader solution space \cite{yin2025dynamic} and conduct iterative self-refinement \cite{cui2025process}. Furthermore, PRMs can be leveraged to train LLMs by providing granular reward signals tailored to intermediate reasoning steps.

\begin{figure*}[htbp]
    \begin{minipage}[b]{0.48\textwidth}
        \centering
        \includegraphics[width=\linewidth]{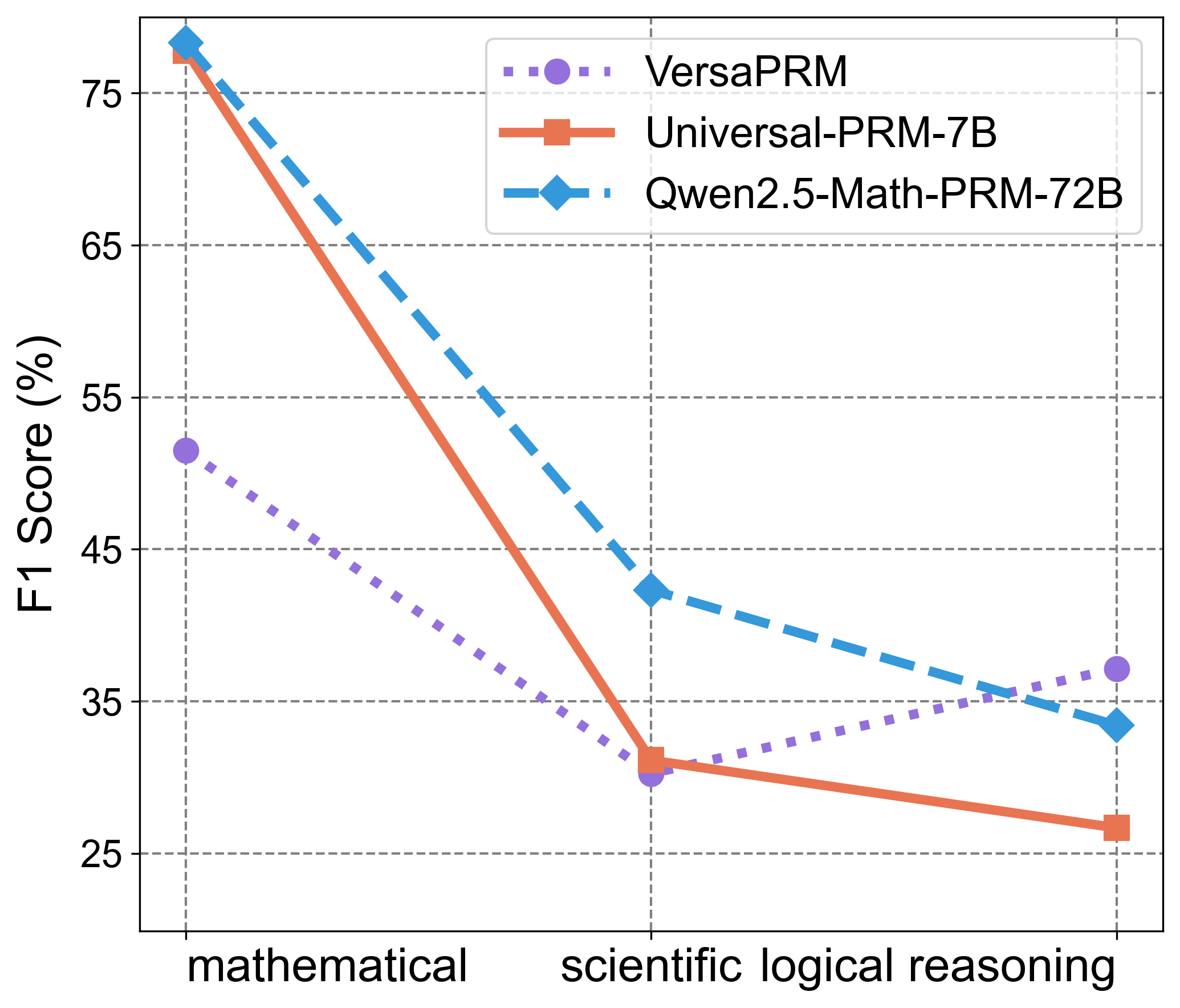} 
        \caption{Compared to mathematical reasoning, PRMs' error detection performance is obviously lower on the domains of logical reasoning and scientific reasoning.}
        \label{fig:intro}
    \end{minipage}
    \hfill
    \begin{minipage}[b]{0.48\textwidth}
        \centering
        \includegraphics[width=\linewidth]{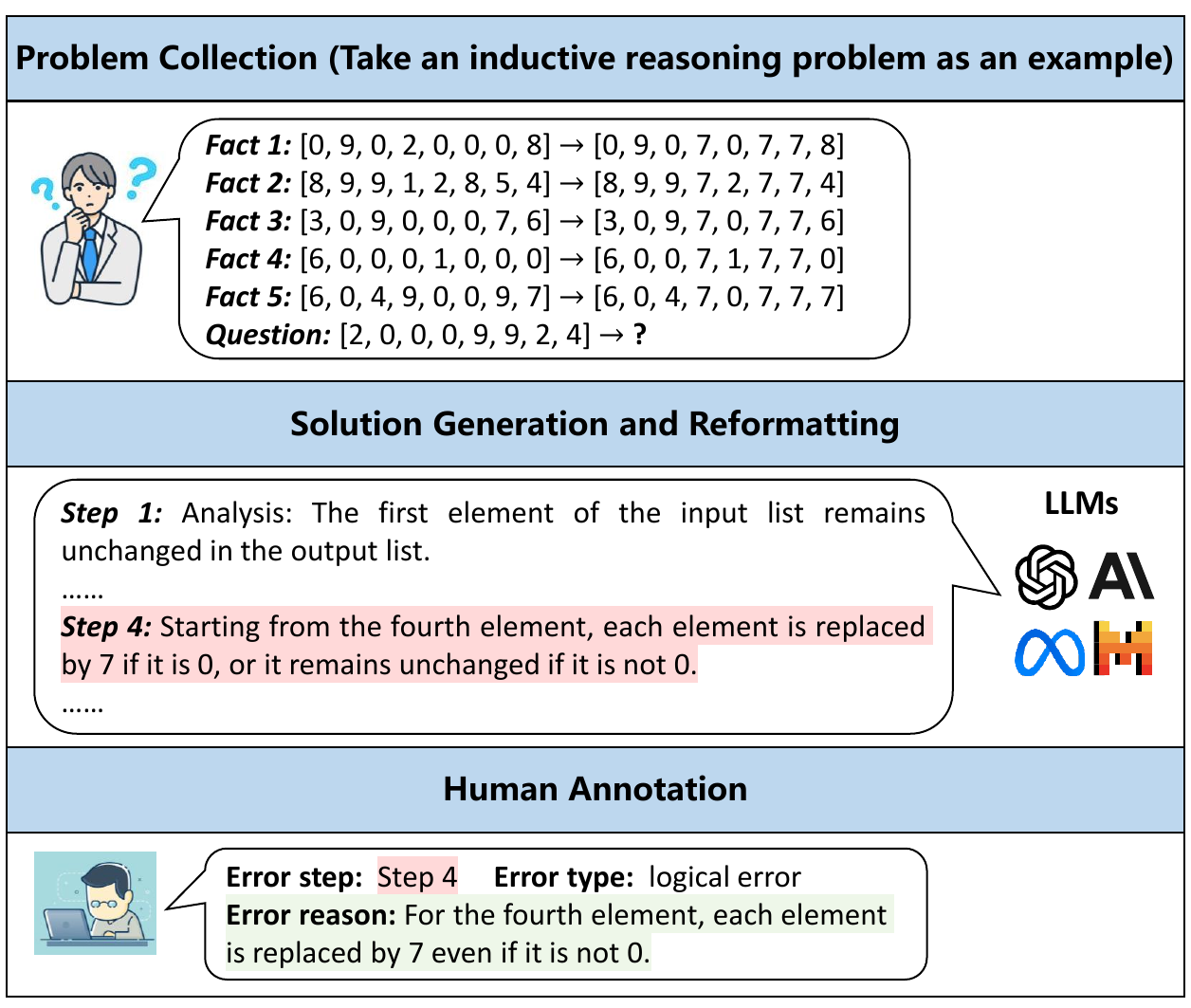}
        \caption{This figure presents an example to briefly illustrate the process of constructing LSR-Ben, including problem collection, solution generation and reformatting, human annotation.}
        \label{fig:datacase}
    \end{minipage}
\end{figure*}

To replicate the success in the field of mathematical reasoning, previous studies have developed some PRMs targeting broader reasoning scenarios such as VersaPRM \cite{zeng2025versaprm} and OpenPRM \cite{zhang2025openprm}. However, a critical limitation across this line of research is that existing benchmarks for PRMs \cite{zheng2025processbench,song-etal-2025-prmbench,li2025socratic} are exclusively tailored to the mathematical domain. In fact, there exists a substantial performance gap between these PRMs’ error-identification capability in the mathematical domain and that in other reasoning domains. As shown in Figure~\ref{fig:intro}, our experimental results demonstrate that the error detection capability of PRMs is far inferior in scientific and logical reasoning domains compared to that in the mathematical domain. This limitation consequently confines the evaluation of PRMs to inefficient and indirect methods such as best-of-N sampling \cite{stiennon2020learning} in reasoning domains beyond math.

To evaluate PRMs across a broader range of reasoning scenarios, we propose a \textbf{L}ogical and \textbf{S}cientific \textbf{R}easoning \textbf{Ben}chmark (LSR-Ben), which is specifically designed to assess PRMs' ability to identify the erroneous steps in broader reasoning scenarios. Our benchmark includes 3,595 data instances spreading across nine subdomains, whose quality is ensured
by professional annotators and the double cross-validation process. We prioritize several principles when designing this benchmark:
\begin{itemize}[leftmargin=*]
\item \textbf{Broader Reasoning Types}: LSR-Ben covers two major reasoning categories (scientific and logical reasoning) along with nine subdomains. This enables a broader evaluation of PRMs in identifying reasoning errors across logical and scientific reasoning domains.

\item \textbf{Wide Solution Coverage}: PRMs can be applied to a wide range of LLMs in practical scenarios. However, the distribution of reasoning processes generated by different LLMs tends to be inconsistent (experimental verifications are provided in Appendix~\ref{sec:coverage}). To ensure a more comprehensive distribution coverage of reasoning processes in the PRM benchmark, we select a diverse set of both open-source and closed-source LLMs to generate reasoning processes for annotation purposes.

\item \textbf{Supports the analysis of different error types}: Beyond erroneous reasoning steps, corresponding error categories are also systematically annotated. This enables us to evaluate the weakness of PRMs in identifying reasoning steps associated with specific error types.
\end{itemize}

We perform an extensive evaluation on LSR-Ben, encompassing two categories of models: PRMs and general-purpose LLMs. For PRMs, we include 13 open-source PRMs to assess step-wise correctness of each reasoning process. For LLMs, we prompt 9 open-source and 2 closed-source LLMs like Qwen3 \cite{qwen3technicalreport} and Gemini-3 \cite{comanici2025gemini} to assess each solution step by step. Experimental results show that despite exhibiting strong error detection capabilities within the mathematical domain, existing PRMs still fail to generalize to scientific and logical reasoning domains. In contrast, LLMs demonstrate non-trivial error identification capabilities in these two reasoning domains. Furthermore, we find that LLMs exhibit a tendency toward over-identification of errors compared to PRMs, whereas PRMs exhibit an inherent tendency to overlook errors compared to LLMs. The full dataset can be found in the supplementary materials.

\begin{table*}[h]
\small
\centering
\caption{Comparison between LSR-Ben and other benchmarks related to reasoning process assessment. 
``Solution Coverage'' denotes the distribution coverage of the input reasoning processes in our benchmarks. The greater the number and diversity of LLMs employed for solution generation, the greater the magnitude of the solution coverage.}
\label{tab:benchmark_compare}
\setlength{\tabcolsep}{0.6mm}       
\renewcommand{\arraystretch}{1.0} 
\begin{tabular}{l c c c c c c c} 
\toprule
& \thead{Domain} & \thead{Step\\Annotation?} & \thead{Solution Generator\\Numbers \& Types} & 
\thead{Solution\\Coverage} & \thead{Annotator} & \thead{Avg\\Steps} & \thead{Error Type\\
Detection?} \\
\midrule
CriticBench      & General & \xmark & 8 ; 3 & \starm & LLM + Human & -- & No \\
MathCheck        & Math    & \cmark & 1 ; 1 & \starl & LLM & 6.4 & No \\
PRMBench         & Math    & \cmark & 2 ; 2 & \starl & Human & 13.4 & Yes \\
ProcessBench     & Math    & \cmark & 12 ; 2 & \starm & Human & 7.1 & No \\
Socratic         & Math    & \cmark & 1 ; 1 & \starl & LLM+Human & 8.7 & Yes \\
\midrule
LSR-Ben          & Logical;Science  & \cmark & 15 ; 8 & \starh & Human & 13.7 & Yes \\
\bottomrule
\end{tabular}
\end{table*}

\section{Related Works}

Process reward models (PRMs) have exhibited considerable potential to enhance process-level reasoning accuracy and long-process reasoning abilities \cite{lightman2023let,zhang2024entropy,zhang-etal-2025-process} by scaling test-time computation. However, PRMs are not always accurate in assessing reasoning processes, highlighting the necessity of proposing evaluation benchmarks for PRMs. While some benchmarks such as CriticBench \cite{lin2024criticbench} exist, they lack step-level correctness annotations so that they cannot be utilized to assess PRMs at the step level.

To address this gap, MathCheck \cite{zhouyour} synthesizes solutions containing erroneous steps. However, the correctness of each reasoning step is also labeled by LLMs, so the resulting annotations suffer from inherent inaccuracies and systemic biases \cite{stolwijk2025generative}.  
To deal with this problem, ProcessBench \cite{zheng2025processbench} generates a set of candidate solutions (i.e., reasoning processes) and employs experts to annotate the first erroneous step in the reasoning process, or to label the reasoning process as error-free. Concurrent work PRMBench \cite{song-etal-2025-prmbench} employs LLMs to synthesize erroneous reasoning steps targeted at a predefined set of error types, followed by manual validation to verify whether these steps conform to the targeted error categories. Furthermore, to evaluate PRMs systematically under six reasoning patterns proposed by the ancient Greek philosopher Socrates, Socratic-PRMBench \cite{li2025socratic} is constructed by  training a specialized Socratic reasoning model to generate Socratic Reasoning processes.  Subsequently, the validity of the reasoning steps is verified through a combined approach of human evaluation and LLM-assisted assessment.
However, all these benchmarks are exclusively tailored to the domain of mathematics, failing to evaluate PRMs in broader reasoning scenarios.

LSR-Ben is distinguished from prior benchmarks or datasets by three key aspects, as highlighted
in Table~\ref{tab:benchmark_compare}. 
First, rather than being confined to the mathematical domain, LSR-Ben primarily encompasses scientific and logical reasoning scenarios, which enables a broader evaluation of PRMs' ability to identify errors.
Second, LSR-Ben leverages a wider variety of LLMs to generate reasoning processes for annotation, thereby ensuring that the distribution of inputs employed in static evaluation can cover a greater spectrum of those encountered in real-world application scenarios.
Third, LSR-Ben contains distinct error types as well as the specific reasons for each erroneous step. This enables us to evaluate the limitations of PRMs and LLMs in identifying reasoning steps associated with specific error types, which can facilitate the development of more advanced PRMs. 

\section{Benchmark Construction}

\subsection{Task Definition}
When a subsequent step in a reasoning process invokes the conclusion derived from a preceding erroneous step, it becomes difficult to definitively determine whether the latter step itself is flawed \cite{lightman2023let}. As a result, following \citet{zheng2025processbench}, LSR-Ben only requires models to either identify the earliest-occurring erroneous step or conclude that all the reasoning steps are correct. Formally, given a reasoning problem $P$ and its step-by-step solution $S = \{s_1, \dots, s_{n}\}$, the task is to output an index $i \in \{-1, 1, ..., n\}$, where $i = -1$ indicates that all steps are correct, and $i \ge 1$ indicates that the earliest error occurs at step $s_i$.

\subsection{Data Curation}
As shown in Figure~\ref{fig:datacase}, we first collect problems encompassing diverse reasoning categories, then leverage LLMs to generate corresponding solutions (i.e., reasoning processes), and finally employ domain experts and crowd-sourced workers to annotate these solutions.

\noindent\textbf{Reasoning Categories} Our benchmark targets two reasoning categories: scientific reasoning and logical reasoning. For scientific reasoning, four subdomains—physics, chemistry, biology, and computer science—are specifically incorporated into the benchmark construction process. As for logical reasoning, it is divided into five subdomains—deductive reasoning, inductive reasoning, abductive reasoning, analogical reasoning, and mixed-form reasoning.

\noindent\textbf{Error Types} We classify reasoning errors into five main categories: knowledge-based errors, factual errors, computational errors, logical errors, and others. \textbf{knowledge-based errors} refer to the employment of extra domain-specific knowledge, commonsense knowledge, or world knowledge that is incorrect during the reasoning process.
\textbf{Factual errors} denote the adoption of facts that contradict the known information in the reasoning steps. For example, if the problem states that John resides in New York, but a reasoning step claims that `it is known that John resides in London', this constitutes a factual error.
\textbf{Logical errors} arise when the conclusion in a reasoning step cannot be logically derived or inferred from the available information.
\textbf{Computational errors} occur when mathematical miscalculations are made in the course of the reasoning process.

\noindent\textbf{Problem Collection} For scientific reasoning (including physics, chemistry, biology, and computer science), we collect problems from the widely used benchmark MMLU-Pro \cite{wang2024mmlu}, which comprehensively assesses LLMs across diverse academic and professional domains. For logical reasoning domains, we curate problem instances from diverse datasets corresponding to distinct subdomains. Specifically, we select FOLIO \cite{han2024folio}, MIRAGE \cite{limirage}, CauseLogics \cite{he2024causejudger}, Analobench \cite{ye2024analobench}, and LogiQA2.0 \cite{liu2023logiqa} for deductive, inductive, abductive, analogical, and mixed-form logical reasoning, respectively. Detailed information about these logical reasoning datasets can be found in Appendix~\ref{sec:datasetdetails}.

\noindent\textbf{Solution Generation and Reformatting} Each instance in our benchmark is a problem–solution pair. To achieve a wider distribution coverage of solutions, we select 11 open-source and 4 closed-source LLMs to generate reasoning processes for annotation purposes. Specifically, we select a suite of representative closed-source and open-source models including Claude \cite{claude}, LLaMA \cite{dubey2024llama}, and Qwen-series models \cite{qwen3technicalreport}, etc. This includes a wide range of model scales, types, and downstream task performance, leading to the high diversity of reasoning processes. More details about the solution generation process are shown in  Appendix~\ref{sec:generatorstatistic}. 

Owing to the diversity of solution generators, the step granularity of the generated solutions exhibits notable inconsistency. Specifically, certain solutions consist of concise yet logically incomplete steps, whereas others incorporate protracted paragraphs that amalgamate multiple logical components. This non-uniformity in step granularity (coupled with the potential for inappropriate step segmentation) hinders the standardization of human annotation criteria.

To tackle this issue, we employ a solution reformatting process to standardize the granularity of reasoning steps, whereby the segmented fragments can be better aligned with logically coherent and appropriately granular reasoning steps. Specifically, we prompt Qwen3-32B \cite{qwen3technicalreport} to insert double line breaks (i.e., segment solutions) while keeping the original solution content unchanged, thereby achieving a structured re-segmentation of the solution text.

Manual analysis for 360 samples finds that this solution reformatting process effectively unifies the granularity of reasoning steps and diminishes the occurrence of logically incomplete steps. We also observe that a small number of solution contents are modified (<1\%) during the process of solution reformatting, so we exclude all solutions whose final answers are changed after reformatting (even if minor alterations to the reasoning process may not undermine the validity of benchmark construction). An illustrative example of solution reformatting is provided in Figure~\ref{fig:reformat_case}.

\begin{table*}[h]
\small
\centering
\caption{Aggregated statistics for all sub-domains (CompSci is short for computer science). Columns represent the statistics for each subfield and the total benchmark without distinguishing between correct and incorrect answers. More detailed statistics can be found in Appendix~\ref{sec:datasetstatistic}. Each data entry is annotated by three different annotators, and only those annotations with a majority consensus were retained for our benchmark. ``\% n/3 agreement” denotes the proportion of samples where the n-annotator agreement is achieved for 3 annotators, so (\% 2/3) + (\% 3/3) = 100\%.}
\label{tab:aggregated_stats_total}
\setlength{\tabcolsep}{0.35mm}       
\renewcommand{\arraystretch}{1.1}
\begin{tabular}{l c c c c c c c c c c}
\toprule
& \multicolumn{4}{c}{\textbf{Scientific Reasoning}} & \multicolumn{5}{c}{\textbf{Logical Reasoning}} & \multicolumn{1}{c}{\multirow{2}{*}[-1ex]{\textbf{Total}}}\\
\cmidrule(lr){2-5} \cmidrule(lr){6-10} & CompSci & Physics & Biology & Chemistry & Deduct & Induct & Analogical & Abduct & Mix \\
\midrule
\textbf{No. Samples} & 400 & 392 & 399 & 391 & 410 & 400 & 402 & 402 & 399 & 3595 \\ 
\midrule
\textbf{Avg. Steps} & 13.4 & 12.6 & 13.7 & 13.4 & 21.1 & 10.7 & 16.6 & 13.3 & 8.0 & 13.7 \\
\midrule
\% 2/3 agreement & 19.0\% & 12.8\% & 23.8\% & 17.6\% &23.8\% & 7.9\% & 20.9\% & 18.9\% & 28.0\% &23.6\% \\
\% 3/3 agreement & 81.0\% & 87.2\% & 76.2\% & 82.4\% &76.2\% & 92.1\% & 79.1\% & 81.1\% & 72.0\% &76.4\% \\
\midrule
\multicolumn{11}{c}{\textit{\textbf{Error Type Distribution}}} \\
\midrule
Knowledge-based errors & 
27.1\% & 19.6\% & 43.5\% & 13.0\% & 0.0\% & 0.0\% & 0.0\% & 0.0\% & 5.5\% & 12.1\% \\

Factual errors & 
21.1\% & 12.3\% & 5.8\% & 9.7\% & 13.3\% & 10.9\% & 5.6\% & 5.5\% & 0.5\% & 9.4\% \\

Computational errors & 
3.0\% & 18.4\% & 4.7\% & 46.5\% & 0.0\% & 6.9\% & 0.0\% & 0.0\% & 0.0\% & 8.8\% \\

Logical errors & 
39.2\% & 44.1\% & 26.2\% & 28.6\% & 85.7\% & 80.7\% & 93.9\% & 94.5\% & 92.5\% & 65.1\% \\

Others & 
9.5\% & 5.6\% & 19.9\% & 2.2\% & 1.0\% & 1.5\% & 0.5\% & 0.0\% & 1.5\% & 4.6\% \\

\bottomrule
\end{tabular}
\end{table*}

\subsection{Human Annotation} 
To ensure a balanced distribution between erroneous and correct solutions, we first use Qwen3-32B to verify the correctness of final answers for each solution. Then, during the annotation process, we dynamically sample solutions with either correct or incorrect final answers, so as to ensure that the proportion of data with valid reasoning processes and those with flawed ones is as balanced as possible in the final dataset.

For the four subdomains of scientific reasoning, we recruited human experts with a bachelor’s degree in the corresponding major, all of whom were required to pass a mandatory competency assessment and annotation training. For the five subdomains of logical reasoning, we engaged the services of a professional annotation company, which meticulously recruited annotators with a bachelor’s degree. All these annotators were also required to pass a mandatory competency assessment and annotation training prior to commencing their tasks. To further ensure annotation quality, a dedicated meta-controller was assigned to conduct rigorous quality inspections across all subdomains.

During the initial annotation process, three distinct annotators were tasked with identifying the first erroneous step in the reasoning process and labeling its error type (if any errors existed in the reasoning process). Additionally, they were also tasked with providing specific error causes for the erroneous steps to facilitate the subsequent double cross-validation process. 
Annotators were permitted to skip a data sample if they deemed their domain knowledge insufficient to complete the annotation for this sample.
If any data sample was annotated by only one annotator (this situation is extremely rare in the annotation process), we skip this data. And we employ another annotator to label the data if there are only two annotation results for one data. 
When the initial three annotators failed to reach a full consensus, a double cross-validation procedure was initiated. In the first round of this procedure, each annotator was provided with the conflicting annotations and corresponding error causes submitted by other annotators. They could either agree with the alternative annotations or provide justifications to refute them. If a full consensus was not reached after the first round, annotators were required to review the refutations from their peers and further determine whether to revise their previous annotations in the second round.
If a majority consensus could still not be achieved after the double cross-validation process (annotation distribution of (1,1,1)), the corresponding solution was discarded from the dataset. Additionally, we also excluded solutions whose final answer was incorrect (according to the reference answer) but the manual annotation of the solution was deemed correct. These resulted in an overall discard rate of approximately 15\% throughout the entire annotation process.

\subsection{Benchmark Statistics}

The resulting LSR-Ben includes nine subsets, consisting of 3,595 test cases in total. Benchmark statistics are shown in Table~\ref{tab:aggregated_stats_total}, the more statistic details are shown in Table~\ref{tab:science_domains} and Table~\ref{tab:logical_domains} (in Appendix~\ref{sec:datasetstatistic}). From Table~\ref{tab:aggregated_stats_total}, it can be found that most data entries reached perfect consensus among the three annotators, indicating the efficacy of our double cross-validation process in ensuring the high quality of the annotated dataset. Furthermore, we observe that current LLMs are prone to all types of errors in the domain of scientific reasoning, whereas in the realm of logical reasoning, their errors are predominantly of a logical error. This is attributed to the design principle of the specific dataset employed for collecting problems. For instance, FOLIO \cite{han2024folio} deliberately excludes any extra knowledge required for solving the problem during its development, to avoid the impact of knowledge deficiency on the deductive reasoning ability of the evaluation model.

\section{Evaluation}

\subsection{Experimental Setup}
\label{sec:setup}
For the evaluation experiments on LSR-Ben, two categories of models are considered: process reward models (PRMs) and general-purpose large language models (LLMs).

\noindent\textbf{Evaluation Metrics} Following \citet{zheng2025processbench}, when conducting evaluations on each subset of the LSR-Ben dataset, we first compute the accuracies for erroneous and correct samples separately, and then derive the harmonic mean of these two accuracy metrics as the \textbf{F1 score} for each LSR-Ben's subset. For the comparison of different models, we prioritize the F1 score as the core evaluation metric, since it effectively balances the trade-off between being overly critical (i.e., over-identifying errors) and being incapable of identifying errors (i.e., failing to detect errors). In addition to LSR-Ben, we further carry out a set of comparative experiments on ProcessBench \cite{zheng2025processbench}, a math-oriented benchmark.

\noindent\textbf{Process Reward Models} Our evaluation includes two categories of open-source PRMs: mathematics-oriented PRMs and non-mathematics-oriented PRMs. To the best of our knowledge, VersaPRM \cite{zengversaprm} and Qwen2.5-PRM-7B-PDDL \cite{pisano2026process} is the only two open-source PRMs that are exclusively trained on other reasoning domains rather than mathematical reasoning. For other open-source PRMs, their training data is confined solely to the field of mathematics. Specifically, open-source mathematics-oriented PRMs include Universal-PRM-7B \cite{tan2025aurora}, Qwen2.5-Math-SCAN-Pro-7B \cite{dingscan}, Qwen2.5-Math-PRM-7B, Qwen2.5-Math-PRM-72B \cite{prmlessons}, Qwen2.5-Math-PRM800K-7B \cite{zheng2025processbench}, ReasonEval-7B, ReasonEval-34B \cite{xia2025evaluating}, FoVer-Qwen2.5-7B \cite{kamoi2026efficient}, and three Llemma-series PRMs \cite{sun2024easy}.

For most PRMs, we can directly extract the earliest erroneous step from their correctness predictions for each reasoning step. However, for PRMs that produce scalar scores for each reasoning step (including Universal-PRM-7B, VersaPRM, and three Llemma-series PRMs), the correctness of reasoning steps cannot be directly derived. As a result, we first transform these scalar scores into binary correctness predictions according to the threshold (a reasoning step is deemed incorrect if its score is below the threshold), and then extract the first erroneous step. To determine the transformation threshold, we first randomly sample 40 (approximately 10\% of the data within each subdomain) label-balanced data (20 with correct reasoning processes and 20 with flawed ones) from each subdomain to construct the domain-specific dev set (when evaluating these PRMs, samples in the dev set are excluded from the benchmark to avoid data leakage). Subsequently, the threshold is determined as the value that yields the highest F1 score on the corresponding domain’s dev set. 

\noindent\textbf{Large Language Models} For LLMs, we prompt them to provide feedback and critique to model-generated solutions. Specifically, LLMs are instructed to output the index corresponding to the reasoning step where the first error arises. We show in Figure~\ref{fig:prompt_template} the prompt template we implement for our evaluation. 

Our experiments include the widely-used DeepSeek-V3.2 \cite{liu2025deepseek}, Llama-3.3-70B-Instruct \cite{grattafiori2024llama}, Kimi-K2-Instruct-0905 \cite{team2025kimi}, Gemma-3-12b-it, Gemma-3-27b-it \cite{team2025gemma}, and Qwen3-series open-source models \cite{qwen3technicalreport}. For closed-source LLMs, we select Gemini-3.8-flash \cite{comanici2025gemini} and GPT-5.5-2026-04-24 \cite{singh2025openai} for evaluation. For open-sourced LLMs, we report the mean performance across three runs. For closed-source LLMs, we report the results under single sampling considering the cost of API. By default, we disable the thinking behavior of LLMs (e.g., explicitly setting enable\_thinking=False for Qwen3-series models).

\begin{table*}[t]
\small
\centering
\caption{Evaluation results of PRMs and LLMs on ProcessBench (abbreviated as PB in the table) and LSR-Ben. Specifically, we report F1 score of the respective accuracies on erroneous and correct samples for each subdomain. Additionally, we also report the average (abbreviated as ave in the table) of the F1 score on nine subdomains. Bio, phys, chem, compsci are short for biology, physics, chemistry and computer science, respectively. In this figure, the best and the second-best results are indicated by bold and underline for PRMs and LLMs, respectively. VersaPRM is italicized to indicate that its training data is not confined to the field of mathematics.}
\label{tab:prm_results}
\setlength{\tabcolsep}{0.5mm}{
    \begin{tabular}{l c c c c c c c c c c c c} 
    \toprule
    \multicolumn{1}{l}{\multirow{3}{*}[-1ex]{\textbf{Model}}} & \multicolumn{1}{c}{\multirow{3}{*}[-1ex]{\textbf{PB}}} & \multicolumn{9}{c}{\textbf{LSR-Ben}} & \\
    \cmidrule(lr){3-12} 
    & & \multicolumn{4}{c}{\textbf{Scientific Reasoning}} & \multicolumn{5}{c}{\textbf{Logical Reasoning}} & \multicolumn{1}{c}{\multirow{2}{*}[-1ex]{\textbf{Ave}}}  \\
    \cmidrule(lr){3-6} \cmidrule(lr){7-11} 
    &Math & Bio & Phys & Chem & CompSci & Abduct & Analogical & Mix & Deduct & Induct &  \\
    \midrule
    \multicolumn{12}{c}{\textit{\textbf{Process Reward Models}}} \\
     \midrule
    ReasonEval-7B & 40.8 & 14.4 & 24.5 & 27.3 & 17.5 & 6.7 & 20.4 & 12.7 & 11.3 & 22.4 & 17.5 \\
    ReasonEval-34B & 51.9 & 18.2 & 22.1 & 23.7 & 25.1 & \underline{25.2} & 3.2 & 8.1 & 21.0 & 2.9 & 16.6 \\
    Llemma-MetaMath-7B & 34.8 & 13.6 & 18.4 & 22.8 & 18.1 & 8.3 & 6.1 & 2.0 & 11.4 & 18.1 & 13.2 \\
    Llemma-PRM800K-7B & 41.9 & 10.4 & 17.3 & 21.5 & 13.2 & 13.2 & 18.8 & 20.1 & 13.1 & 17.1 & 16.1 \\
    Llemma-OPRM-7B & 45.9 & 14.8 & 5.4 & 11.0 & 12.1 & 12.1 & 7.9 & 5.5 & 12.9 & 1.9 & 9.3 \\
    Qwen2.5-Math-SCAN-Pro-7B & 64.7 & 20.3 & 31.8 & 27.8 & 15.3 & 14.2 & 34.2 & 8.6 & 13.1 & 12.7 & 19.8 \\
    Qwen2.5-Math-PRM800K-7B & 56.5 & 17.1 & 22.3 & 27.3 & 10.4 & 2.0 & 25.4 & 17.2 & 9.9 & \underline{36.9} & 18.7 \\
    FoVer-Qwen2.5-7B & 51.5 & 24.2 & 25.9 & 27.1 & 18.7 & 12.8 & 26.6 & 24.2 & 22.7 & 29.5 & 23.5 \\
    Universal-PRM-7B & \underline{77.8} & 24.4 & 34.4 & \underline{36.9} & \underline{28.8} & 16.0 & 34.5 & \textbf{43.2} & 22.1 & 17.6 & 28.7 \\
    Qwen2.5-Math-PRM-7B & 73.5 & 22.9 & 34.6 & 32.5 & 20.1 & 3.9 & 40.7 & 27.7 & 22.2 & 33.9 & 26.5 \\
    Qwen2.5-Math-PRM-72B & \textbf{78.3} & \underline{33.6} & \textbf{48.7} & \textbf{40.8} & \textbf{46.2} & 8.5 & \underline{49.4} & 36.7 & \textbf{36.7} & 35.8 & \textbf{37.4} \\
    \textit{Qwen2.5-PRM-7B-PDDL} & 73.6 & 23.7 & \underline{37.4} & 34.6 & 27.6 & 7.6 & 44.4 & 34.2 & \underline{23.9} & 25.6 & 28.8 \\
    \textit{VersaPRM (8B)} & 51.5 & \textbf{44.3} & 31.0 & 22.4 & 23.3 & \textbf{40.0} & \textbf{54.2} & \underline{41.0} & 11.3 & \textbf{39.3} & \underline{34.1} \\
    \midrule
    \multicolumn{12}{c}{\textit{\textbf{Open-source language models, prompted to identify reasoning errors}}} \\
    \midrule
    Llama-3.3-70B-Instruct & 52.6 & 27.7 & 22.2 & 16.7 & 28.9 & 8.6 & 26.9 & 18.9 & 22.0 & 28.8 & 22.3 \\
    Qwen3-4B & 54.1 & 35.1 & 26.9 & 22.0 & 27.5 & 19.0 & 37.9 & 39.4 & 24.4 & 31.1 & 29.3 \\
    Qwen3-8B & 59.3 & 37.5 & 31.1 & 25.6 & 35.7 & 19.9 & 52.1 & 41.9 & 29.9 & 37.0 & 34.5 \\
    Qwen3-14B & 59.4 & 38.1 & 38.9 & 29.0 & 29.3 & 24.7 & 50.0 & 39.6 & 28.2 & 32.2 & 34.4 \\
    Qwen3-32B & 65.9 & 41.7 & 40.1 & 26.3 & 35.0 & 22.0 & \underline{53.5} & 45.6 & 34.0 & 32.1 & 36.7 \\
    Gemma-3-12B-it & 55.1 & 36.4 & 28.3 & 27.1 & 29.1 & 10.2 & 20.6 & 30.7 & 21.7 & 23.1 & 25.2 \\
    Gemma-3-27B-it & 60.7 & 44.8 & 35.6 & 25.4 & 31.2 & 33.1 & 49.1 & 36.4 & 21.6 & 34.9 & 34.7 \\
    Kimi-K2-Instruct-0905 & 75.2 & 47.4 & 51.7 & 42.9 & 51.3 & 49.9 & 45.7 & 47.9 & 38.5 & 66.5 & 49.1 \\
    DeepSeek-v3.2 & \underline{79.8} & \underline{49.9} & \underline{61.5} & \underline{50.3} & \underline{58.5} & \textbf{71.4} & 42.2 & 52.3 & \underline{44.9} & \underline{67.5} & \underline{55.4} \\
    \midrule
    \multicolumn{12}{c}{\textit{\textbf{Proprietary language models, prompted to identify reasoning errors}}} \\
    \midrule
    GPT-5.5-2026-04-24 & 72.7 & 47.3 & 52.7 & 43.7 & 57.0 & 54.4 & 38.6 & \underline{60.6} & 35.2 & 61.1 & 50.1 \\
    Gemini-3.8-flash & \textbf{83.4} & \textbf{61.6} & \textbf{67.6} & \textbf{55.3} & \textbf{69.4} & \underline{71.3} & \textbf{55.2} & \textbf{69.0} & \textbf{48.3} & \textbf{76.0} & \textbf{63.8} \\
    \bottomrule
    \end{tabular}
}
\end{table*}

\subsection{Evaluations of PRMs and LLMs} 
The main experimental results for process reward models and large language models are presented in Table~\ref{tab:prm_results}, with comprehensive details provided in Appendix~\ref{sec:detailevaluationresults}. From these results, we observe that: 

\textbf{(1)} Comparing the performance of PRMs on ProcessBench and LSR-Ben, it is evident that the performance on LSR-Ben is much lower (<30\% in general) than that on ProcessBench, which demonstrates that PRM's ability to determine the correctness of the intermediate steps in logical and scientific reasoning domains is significantly inferior to that exhibited in the mathematical reasoning domain. This highlights the importance of LSR-Ben for developing more advanced PRMs tailored to other reasoning domains rather than only focusing on mathematical reasoning.

\textbf{(2)} Compared to math-oriented PRMs, the non-math-oriented VersaPRM and Qwen2.5-PRM-7B-PDDL achieve superior performance at the same model scale on LSR-Ben. However, their performance is still far below the expectations required in practical applications, demonstrating the necessity of developing PRMs with enhanced efficacy in broader reasoning domains.


\textbf{(3)} Compared to mathematical reasoning, the performance gap between the best LLM (Gemini-3.8-flash) and the best PRM (Qwen2.5-Math-PRM-72B) is larger in logical and scientific reasoning domains, which indicates that PRMs possess considerable potential to achieve superior performance in these two reasoning domains. Meanwhile, it also highlights the importance of developing more advanced methods for training PRMs tailored to broader reasoning scenarios.

\textbf{(4)} In general, both PRMs and LLMs exhibit a markedly lower accuracy on erroneous samples relative to correct samples (as shown in Table~\ref{tab:gr_ben_part1}, \ref{tab:gr_ben_part2} and \ref{tab:gr_ben_part3}), which demonstrates that these models tend to adopt a conservative decision-making stance and thus suffer from the under-identification of reasoning errors. This finding sheds light on the necessity of placing greater emphasis on enhancing the capability of PRMs to detect erroneous reasoning steps during their design and development.

\begin{table*}[t]
  \small
  \centering
  \caption{Breakdown of LLM and PRM predictions on erroneous reasoning processes by error type.  FP\%: the proportion of samples in which the model identifies erroneous reasoning steps prior to the first ground-truth flawed reasoning step, relative to the total number of samples involving flawed reasoning steps. FN\%: the proportion of samples in which the model correctly recognizes all valid reasoning steps preceding the first ground-truth erroneous reasoning step, yet fails to detect this first ground-truth erroneous reasoning step among the data involving flawed reasoning steps.}
  \setlength{\tabcolsep}{0.9mm}{
  \begin{tabular}{l c c c c c c c c c c c c}
  \midrule
    \multicolumn{1}{l}{\multirow{2}{*}[-1ex]{\textbf{Model}}} & \multicolumn{2}{c}{\textbf{Knowledge}} & \multicolumn{2}{c}{\textbf{Factual}} & \multicolumn{2}{c}{\textbf{Computational}} & \multicolumn{2}{c}{\textbf{Logical}} & \multicolumn{2}{c}{\textbf{Others}} & \multicolumn{2}{c}{\textbf{All}}  \\
    \cmidrule(lr){2-3} \cmidrule(lr){4-5} \cmidrule(lr){6-7} \cmidrule(lr){8-9} \cmidrule(lr){10-11} \cmidrule(lr){12-13}  
    & FP\% & FN\% & FP\% & FN\% & FP\% & FN\% & FP\% & FN\% & FP\% & FN\% & FP\% & FN\% \\
    \midrule
    \multicolumn{13}{c}{\textit{\textbf{Process Reward models}}} \\
    \midrule
    Universal-PRM-7B & 26.1 & 54.1 & 31.9 & 39.8 & 29.1 & 41.9 & 36.1 & 39.3 & 45.7 & 38.3 & 34.4 & 41.2 \\
    Qwen2.5-PRM-7B-PDDL & 14.0 & 72.5 & 21.7 & 53.6 & 19.6 & 56.8 & 20.8 & 57.0 & 28.7 & 57.5 & 20.4 & 58.5 \\
    Qwen2.5-Math-PRM-72B & 7.7 & 72.5 & 18.1 & 48.2 & 20.9 & 52.7 & 16.8 & 56.4 & 24.7 & 45.7 & 16.6 & 56.7 \\
    VersaPRM & 8.7 & 69.6 & 15.1 & 63.3 & 19.6 & 67.6 & 21.6 & 49.4 & 29.6 & 45.7 & 19.6 & 54.4 \\
    \midrule
    \textbf{Average} & 14.1 & 67.2 & 21.7 & 51.2 & 22.3 & 54.7 & 23.8 & 50.5 & 32.2 & 46.8 & 22.7 & 52.7 \\
    \midrule
    \multicolumn{13}{c}{\textit{\textbf{Large Language models}}} \\
    \midrule
    Kimi-K2-Instruct-0905 & 22.2 & 36.2 & 15.1 & 33.7 & 19.6 & 54.7 & 23.6 & 33.1 & 24.7 & 44.4 & 22.4 & 35.8 \\
    DeepSeek-v3.2 & 20.3 & 30.9 & 21.1 & 23.5 & 31.8 & 18.9 & 27.0 & 28.1 & 25.9 & 45.7 & 26.0 & 28.0 \\
    GPT-5.5-2026-04-24 & 34.3 & 15.5 & 39.2 & 10.8 & 34.5 & 43.2 & 44.0 & 16.5 & 47.5 & 16.2 & 41.7 & 18.1 \\
    Gemini-3.8-flash & 18.4 & 32.9 & 19.9 & 20.5 & 25.7 & 28.4 & 25.6 & 22.8 & 28.7 & 26.2 & 24.3 & 24.4 \\
    \midrule
    \textbf{Average} & 23.8 & 28.9 & 23.8 & 22.1 & 27.9 & 36.3 & 30.0 & 25.1 & 31.7 & 33.1 & 28.6 & 26.6 \\
    \bottomrule
  \end{tabular}
  }
  \label{tab:error_type_breakdown}
\end{table*}

\subsection{Error Analysis}
To conduct an in-depth analysis for identifying the specific categories of errors that are more difficult to identify, we calculate the false negative (FN) rate and false positive (FP) rate of the model on data involving flawed reasoning steps. Specifically, FP rate refers to the proportion of samples in which the model identifies erroneous reasoning steps prior to the first ground-truth flawed reasoning step, relative to the total number of samples involving flawed reasoning steps. FN rate refers to the proportion of samples in which the model correctly recognizes all valid reasoning steps preceding the first ground-truth erroneous reasoning step, yet fails to detect this first ground-truth erroneous reasoning step among the data involving flawed reasoning steps. Formally, we denote the number of instances in which the model-predicted error location precedes or lags behind the one corresponding to the standard answers as ${n}_{earlier}$ and ${n}_{miss}$, respectively. The FN and FP rates for the instances corresponding to each error type are defined as:

\setlength{\abovedisplayskip}{-1.5ex}
\begin{align*}
\tag{1}   FN = \frac{n_{miss}}{n_{total}}, FP = \frac{n_{earlier}}{n_{total}} 
\label{eq:1}
\end{align*}
where ${n}_{total}$ is the total number of instances involving flawed reasoning steps corresponding to a specific error type. To derive conclusions conducive to further improving the error detection performance of PRMs and LLMs, we focus primarily on well-performing PRMs and LLMs. Specifically, we conduct experiments on the top four performing PRMs and the top four performing LLMs as presented in Table~\ref{tab:prm_results}, and present the experimental results categorized by error type in Table~\ref{tab:error_type_breakdown}. 

Experimental results show that, on average, PRMs exhibit the weakest capability in identifying knowledge-based errors (according to FN), while LLMs tend to perform worst at detecting computational errors. Furthermore, we find that the FN rate of PRMs is significantly higher than that of LLMs, whereas the FP rate of LLMs is greater than that of PRMs. This indicates that LLMs exhibit a tendency toward over-identification of errors compared to PRMs, whereas PRMs exhibit an inherent tendency to overlook errors compared to LLMs. This reflects a complementary capability relationship, which can be exploited to further enhance the overall error detection performance of PRMs.

\begin{table*}[t]
\small
\centering
\caption{Evaluation results of different thinking modes on ProcessBench (abbreviated as PB in the table) and LSR-Ben for two closed-source and five open-source LLMs. Specifically, we report F1 score of the respective accuracies on erroneous and correct samples for each subdomain. Additionally, we report the average of the F1 score on nine subdomains. Bio, phys, chem, compsci are short for biology, physics, chemistry and computer science, respectively. Fast and Slow refer to deactivating and activating the slow thinking mode, respectively.}
\label{tab:slow}
\setlength{\tabcolsep}{0.4mm}{
    \begin{tabular}{l c *{10}{c} c} 
    \toprule
    \multicolumn{1}{l}{\multirow{3}{*}[-1ex]{\textbf{Model}}} & \multicolumn{1}{c}{\multirow{3}{*}[-1ex]{\textbf{PB}}} & \multicolumn{9}{c}{\textbf{LSR-Ben}} & \\
    \cmidrule(lr){3-12} 
    & & \multicolumn{4}{c}{\textbf{Scientific Reasoning}} & \multicolumn{5}{c}{\textbf{Logical Reasoning}} & \multicolumn{1}{c}{\multirow{2}{*}[-1ex]{\textbf{Average}}} \\
    \cmidrule(lr){3-6} \cmidrule(lr){7-11} 
    &Math & Bio & Phys & Chem & CompSci & Abduct & Analogical & Mix & Deduct & Induct & \\
    \addlinespace[3pt]
    \midrule
    GPT-5.5-2026-04-24 \textit{(Fast)} & 72.7 & 47.3 & 52.7 & 43.7 & 57.0 & 54.4 & 38.6 & 60.6 & 35.2 & 61.1 & 50.1 \\
    GPT-5.5-2026-04-24 \textit{(Slow)} & 80.1 & 58.1 & 67.2 & 52.1 & 62.3 & 69.7 & 40.7 & 63.0 & 43.0 & 73.8 & 58.9 \\
    \midrule
    Gemini-3.8-flash \textit{(Fast)} & 83.4 & 61.6 & 67.6 & 55.3 & 69.4 & 71.3 & \textbf{55.2} & \textbf{69.0} & 48.3 & \textbf{76.0} & 63.8 \\
    Gemini-3.8-flash \textit{(Slow)} & \textbf{84.4} & \textbf{62.8} & \textbf{69.2} & \textbf{56.5} & \textbf{72.0} & 73.0 & 52.5 & 67.7 & \textbf{50.1} & 75.6 & \textbf{64.4} \\
    \midrule
    DeepSeek-v3.2 \textit{(Fast)} & 79.8 & 49.9 & 61.5 & 50.3 & 58.5 & 71.4 & 42.2 & 52.3 & 44.9 & 67.5 & 55.4 \\
    DeepSeek-v3.2 \textit{(Slow)} & 81.0 & 51.0 & 62.2 & 53.6 & 59.5 & \textbf{73.1} & 44.2 & 54.6 & 41.0 & 71.2 & 56.7 \\
    \midrule
    Qwen3-4B \textit{(Fast)} & 54.1 & 35.1 & 26.9 & 22.0 & 27.5 & 19.0 & 37.9 & 39.4 & 24.4 & 31.1 & 29.3 \\
    Qwen3-4B \textit{(Slow)} & 73.6 & 35.9 & 45.1 & 35.8 & 34.8 & 5.8 & 28.4 & 36.5 & 25.0 & 45.6 & 32.5 \\
    \addlinespace[0.5mm]
    Qwen3-8B \textit{(Fast)} & 59.3 & 37.5 & 31.1 & 25.6 & 35.7 & 19.9 & 52.1 & 41.9 & 29.9 & 37.0 & 34.5 \\
    Qwen3-8B \textit{(Slow)} & 75.7 & 39.0 & 46.1 & 36.0 & 38.2 & 13.9 & 49.9 & 47.9 & 36.0 & 50.9 & 39.8 \\
    \addlinespace[0.5mm]
    Qwen3-14B \textit{(Fast)} & 59.4 & 38.1 & 38.9 & 29.0 & 29.3 & 24.7 & 50.0 & 39.6 & 28.2 & 32.2 & 34.4 \\
    Qwen3-14B \textit{(Slow)} & 76.0 & 45.1 & 49.9 & 37.5 & 47.3 & 16.4 & 52.1 & 47.3 & 40.2 & 44.6 & 42.3 \\
    \addlinespace[0.5mm]
    Qwen3-32B \textit{(Fast)} & 65.9 & 41.7 & 40.1 & 26.3 & 35.0 & 22.0 & 53.5 & 45.6 & 34.0 & 32.1 & 36.7 \\
    Qwen3-32B \textit{(Slow)} & 76.5 & 44.0 & 47.9 & 44.0 & 47.6 & 32.9 & 57.3 & 46.2 & 39.4 & 49.1 & 45.4 \\
    \bottomrule
    \end{tabular}
}
\end{table*}

\subsection{Different Thinking Modes Comparison}
Slow thinking is a thoughtful process where LLMs decompose, reflect on, and plan for a problem prior to generating a formal response. This line of thinking mode has been successfully implemented in a variety of LLMs such as Gemini-3.8-flash \cite{comanici2025gemini} and GPT-5.5 \cite{singh2025openai}. To investigate the influence of slow thinking on the error-detection ability of LLMs, we utilize two closed-source and five open-source LLMs for experiments. When conducting experiments of slow thinking, we set the max output tokens to 32,768 in order to avoid premature generation termination, and also set the reasoning effort to `xhigh' and `high' for GPT-5.5 and Gemini-3.8-flash, respectively. 

Experimental results are shown in Table~\ref{tab:slow}, from which we can find that using slow thinking can improve the error identification ability of the same LLM by \textit{thinking} more before judging the correctness of each step. However, even if equipped with slow thinking, the best performance of LLMs is still far from expectations in logical and scientific reasoning domains, which underscores the difficulty level of detecting errors in broader reasoning scenarios. In addition, we observe that slow thinking exhibits suboptimal performance in the domain of analogical reasoning. This can be attributed to the fact that analogical reasoning tasks only require inferential processes grounded in superficial semantic relationships. Consequently, the employment of slow thinking in such tasks tends to give rise to the issue of overthinking. 

\section{Conclusions}
We introduce the LSR-Ben benchmark for measuring the PRMs' ability to identify erroneous steps in broader reasoning scenarios, characterized by its diversity of reasoning types, rigorous human annotation, support for the analysis of error types, and wide solution coverage. Through extensive evaluation with existing PRMs and LLMs, we find that existing PRMs fail to identify errors in logical and scientific reasoning domains. In contrast, LLMs demonstrate non-trivial error identification capabilities in logical and scientific reasoning domains. Further error analysis demonstrates that LLMs exhibit a tendency toward over-identification of errors compared to PRMs, whereas PRMs exhibit an inherent tendency to overlook errors compared to LLMs. 

\subsection*{AI use statement}
In this work, we used generative AI tools for [polishing the paper writing].
We have not used generative AI tools for [other tasks with required disclosure],
and [the rest of the required disclosure tasks] are not applicable to this work.
Additionally, we did not use generative AI tools for [tasks with recommended
disclosure]. We have reviewed all AI-assisted work. [We checked LLM-generated text in this paper]. We take responsibility for the final content of this work,
including text, claims, or artifacts produced with the aid of generative AI.

\subsection*{Ethics statement}
Our work involves human annotators for data annotation. To determine the compensation, we first conducted pilot data annotation. On average, it took participants approximately 10 minutes to annotate one data entry. Therefore, we set the compensation such that each annotator was paid at a rate of 2 US dollars per question, equivalent to an hourly wage of 12 US dollars, which exceeds the national minimum hourly wage in our country.

\subsection*{Reproducibility statement}
We provide the code and data for our experiments in the supplementary material. We also provide a README.md in the supplementary material to providing step-by-step instructions for reproducing our experimental results.

\bibliography{iclr2027_conference}
\bibliographystyle{iclr2027_conference}

\appendix
\section{Appendix}
\begin{figure*}[!h]
    \centering
    \includegraphics[width=0.9\linewidth]{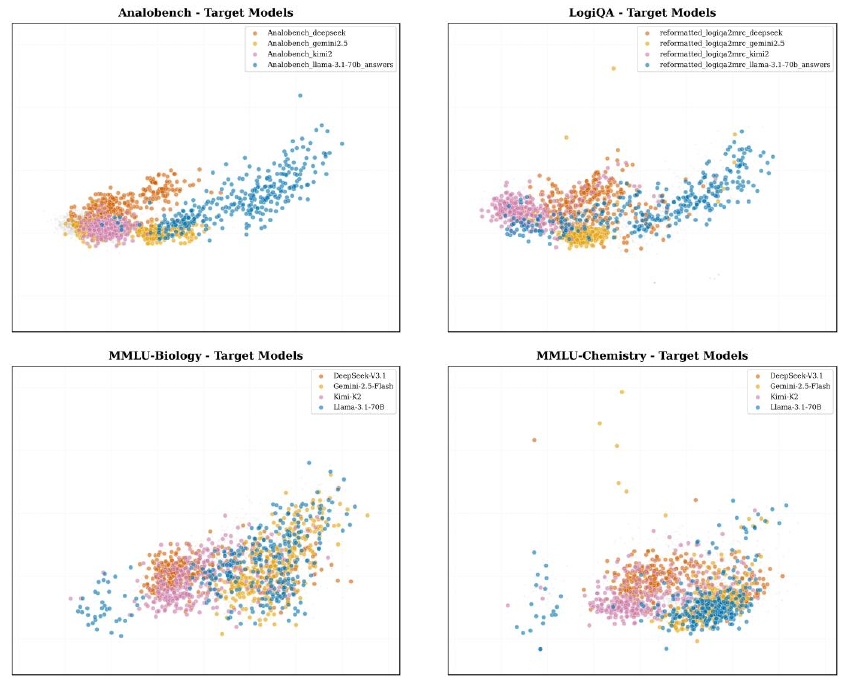}
    \caption{Visualization of solutions generated by different LLMs.}
    \label{fig:coverage}
\end{figure*}

\subsection{Experiments for Solution Coverage}
\label{sec:coverage}

To verify that different LLMs produce solutions with distinct distributions, we use Qwen3-8b to obtain vector representations (using the representation vector of the last token in the model’s final layer) of solutions generated by different LLMs (we also ensure that identical data subsets are employed across all models for solution generation), then apply PCA to reduce their dimensionality to two dimensions for visualization. 

Visualization results are shown in Figure~\ref{fig:coverage}. From these results, we can observe that the distributions of solutions generated by different LLMs exhibit generally substantial discrepancies. As a result, employing a more diverse set of LLMs for solution generation can effectively enhance solution coverage.

\subsection{Dataset Details}
\label{sec:datasetdetails}
\noindent\textbf{FOLIO} is a meticulously designed deductive reasoning corpus grounded in first-order logic. Each problem is rigorously curated to ensure that it can be solved exclusively via first-order logical inference, without recourse to any external or domain-specific knowledge.

\noindent\textbf{MIRAGE} is a synthetic inductive logical reasoning dataset curated by first constructing vector transformation rules, followed by substituting instantiated vectors into these rules to generate factual statements and corresponding question-answer pairs. 

\noindent\textbf{Analobench} is an analogical reasoning dataset. Its task format is defined as follows: given a specific target story, the goal is to select the most analogous candidate from a set of four options. The dataset is divided into two distinct task types: sentence-level analogy and paragraph-level analogy. Specifically, the sentence-level analogy data is carefully curated by human annotators, whereas the paragraph-level analogy data is generated by expanding the original sentence-level analogy data using LLMs. However, since the paragraph-level analogy data lacks manual validation, it suffers from inferior quality. For this reason, we exclusively adopt the sentence-level analogy data for our benchmark construction.

\noindent\textbf{CauseLogics} is a synthetic abductive logical reasoning dataset curated by constructing premises, rules, observed phenomena, and possible causes. The task is to find the root (deepest) cause that accounts for the observed phenomenon (it is ensured that the observed phenomenon can not be inferred by the given premises and rules). 

\noindent\textbf{LogiQA2.0} is a crowdsourced logical reasoning dataset, where each instance may require multiple types of logical reasoning abilities to complete.

\subsection{An Example of Solution Reformatting}
\label{sec:reformat}
\begin{figure*}[!h]
    \centering
    \includegraphics[width=1.0\linewidth]{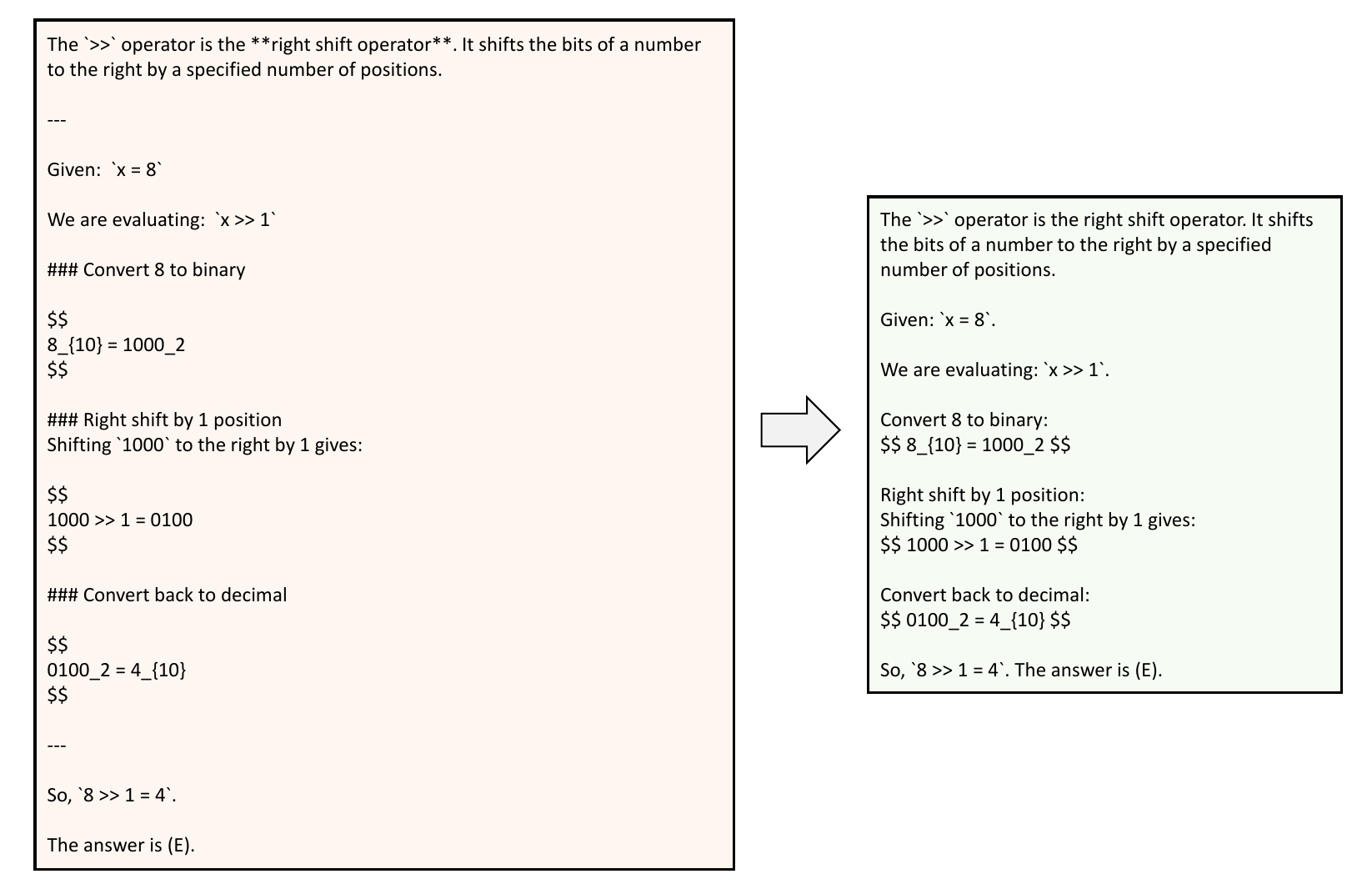}
    \caption{Example of solution reformatting. The left is the original solution and the right is the reformatted one. The problem is `Question: Let x = 8. What is x>>1 in Python 3? Options: A. 5, B. 3, C. 0, D. 8, E. 4, F. 7, G. 1, H. 2, I. 6, J. 16'.}
    \label{fig:reformat_case}
\end{figure*}

\subsection{Detailed Dataset Statistics of LSR-Ben}
\label{sec:datasetstatistic}
\begin{table*}[!h]
\centering
\caption{Detailed Statistics for the four subdomains of scientific reasoning (computer science, physics, biology, chemistry).}
\label{tab:science_domains}
\setlength{\tabcolsep}{1mm}       
\renewcommand{\arraystretch}{1.0} 
\small 
\begin{tabular}{l c c c c c c c c}
\toprule
& \multicolumn{2}{c}{\textbf{Computer science}} & \multicolumn{2}{c}{\textbf{Physics}} & \multicolumn{2}{c}{\textbf{Biology}} & \multicolumn{2}{c}{\textbf{Chemistry}} \\
\cmidrule(lr){2-3} \cmidrule(lr){4-5} \cmidrule(lr){6-7} \cmidrule(lr){8-9} & \textbf{error} & \textbf{correct} & \textbf{error} & \textbf{correct} & \textbf{error} & \textbf{correct} & \textbf{error} & \textbf{correct} \\
\midrule
\textbf{No. Samples} & 199 & 201     & 179 & 213    & 191 & 208     & 185 & 206  \\ 
\midrule
\textbf{Avg. Steps} & 14.3 & 12.5 & 14.3 & 11.1 & 14.4 & 13.2 & 15.0 & 12.0 \\
\midrule
\% 2/3 agreement & 26.1\% & 11.9\% & 20.1\% & 6.6\% & 30.4\% & 17.8\% & 22.7\% & 13.1\% \\
\% 3/3 agreement & 73.9\% & 88.1\% & 79.9\% & 93.4\% & 69.6\% & 82.2\% & 77.3\% & 86.9\% \\
\bottomrule
\end{tabular}
\end{table*}

\begin{table*}[!h]
\small
\centering
\caption{Detailed Statistics for the five subdomains of logical reasoning  (deductive, inductive, analogical, abductive, mixed-form reasoning).}
\label{tab:logical_domains}
\setlength{\tabcolsep}{1mm}       
\renewcommand{\arraystretch}{1.0} 
\footnotesize 
\begin{tabular}{l c c c c c c c c c c}
\toprule
& \multicolumn{2}{c}{\textbf{Deductive}} & \multicolumn{2}{c}{\textbf{Inductive}} & \multicolumn{2}{c}{\textbf{Analogical}} & \multicolumn{2}{c}{\textbf{Abductive}} & \multicolumn{2}{c}{\textbf{Mix}} \\
\cmidrule(lr){2-3} \cmidrule(lr){4-5} \cmidrule(lr){6-7} \cmidrule(lr){8-9} \cmidrule(lr){10-11} & \textbf{error} & \textbf{correct} & \textbf{error} & \textbf{correct} & \textbf{error} & \textbf{correct} & \textbf{error} & \textbf{correct} & \textbf{error} & \textbf{correct} \\
\midrule
\textbf{No. Samples} & 210 & 200 & 202 & 198 & 198 & 204 & 200 & 202 & 199 & 200  \\ 
\midrule
\textbf{Avg. Steps} & 22.8 & 19.4 & 9.6 & 11.7 & 16.1 & 17.0 & 12.9 & 13.6 & 7.7 & 8.4 \\
\midrule
\% 2/3 agreement & 31.3\% & 16.0\% & 12.3\% & 3.5\% & 28.3\% & 13.7\% & 29.0\% & 8.9\% & 36.5\% & 19.5\% \\
\% 3/3 agreement & 68.7\% & 84.0\% & 87.7\% & 96.5\% & 71.7\% & 86.3\% & 71.0\% & 91.1\% & 63.5\% & 80.5\% \\
\bottomrule
\end{tabular}
\end{table*}

\begin{figure*}[!h] 
\subsection{Prompt Template for Large Language Model Evaluation}
\label{sec:eval_prompt}
\vspace{9pt}
    \centering
    \begin{tcolorbox}[colback=white, colframe=black, sharp corners, boxrule=0.8pt, width=\linewidth,
        boxsep=2pt, left=2pt, right=2pt, top=2pt, bottom=2pt]
        \ttfamily 
        \small    
        \renewcommand{\baselinestretch}{0.9}\selectfont 
        \setlength{\parskip}{0pt} 
        \color{black!80} 
        The following is a problem and a solution (split into reasoning steps, enclosed with tags and indexed from 0):
        \par\vspace{\baselineskip}
        [Problem]
        \par\vspace{\baselineskip}
        \textcolor{mainblue}{...(problem)...}
        \par\vspace{\baselineskip}
        [Solution]
        \par\vspace{\baselineskip}
        \textcolor{mainblue}{<reasoning\_step\_1>} \\
        \textcolor{mainblue}{...(reasoning step 1 of solution)...} \\
        \textcolor{mainblue}{</reasoning\_step\_1>}
        \par\vspace{\baselineskip}
        ...
        \par\vspace{\baselineskip}
        \textcolor{mainblue}{<reasoning\_step\_n>} \\
        \textcolor{mainblue}{...(reasoning step n of solution)...} \\
        \textcolor{mainblue}{</reasoning\_step\_n>}
        \par\vspace{\baselineskip}
        Your task is to review and critique the solution step by step. Once you identify an error in a step, return the \textbf{\textcolor{mainred}{index of the reasoning step where the earliest error occurs}}. Otherwise, return the \textbf{\textcolor{mainred}{index of -1}} (which typically denotes "not found").
        \par\vspace{\baselineskip}
        \textbf{\textcolor{mainred}{Please put your final answer (i.e., the index) in \textbackslash boxed\{\}.}}
    \end{tcolorbox}
    \caption{Prompt template for LLM evaluation. The \textbf{\textcolor{mainblue}{blue texts}} indicate the input problem and the solution (split into reasoning steps). The \textbf{\textcolor{mainred}{red texts}} describe the required output content and format.}
    \label{fig:prompt_template}
\end{figure*}

\subsection{More details about the solution generation process}
\label{sec:generatorstatistic}

During solution generation, we kept each LLM’s decoding configuration fixed and used a dataset-specific prompt that was shared by all generators for each dataset. For all the LLMs used for generating solutions, we set temperature to 0.7 and top-p to 0.8. To enhance the diversity of problems, we also strive to ensure that the problem within each instance is unique. However, due to the insufficient number of source problems, a small number of duplicate problems exist within our constructed benchmark (less than 10\%). Furthermore, we artificially ensure that all problems in instances with fully correct reasoning processes are unique, and all problems in instances containing erroneous reasoning steps are also unique. Table~\ref{tab:science_results} and Table~\ref{tab:logical_results} present the breakdown information of LLMs used for LSR-Ben’s solution generation.

\begin{table*}[!h]
\center
\caption{Breakdown statistics of the scientific reasoning domain for LSR-Ben.}
\label{tab:science_results}
\small 
\setlength{\tabcolsep}{2mm} 
\begin{tabular}{l c c c c c c c c}
\toprule
\multirow{2}{*}{\textbf{Model}} & \multicolumn{2}{c}{\textbf{Biology}} & \multicolumn{2}{c}{\textbf{Physics}} & \multicolumn{2}{c}{\textbf{Chemistry}} & \multicolumn{2}{c}{\textbf{CompSci}} \\
\cmidrule{2-3} \cmidrule{4-5} \cmidrule{6-7} \cmidrule{8-9}

& \textbf{Error} & \textbf{Correct} & \textbf{Error} & \textbf{Correct} & \textbf{Error} & \textbf{Correct} & \textbf{Error} & \textbf{Correct} \\
\midrule
Gemma3-12B-IT       & 14 & 14& 21 & 9& 21 & 7& 15 & 12\\
Gemma3-27B-IT       & 9 & 16& 11 & 14& 18 & 10& 19 & 9\\
Qwen3-4B            & 19 & 11& 11 & 16& 15 & 14& 15 & 12\\
Qwen3-8B            & 17 & 12& 12 & 16& 15 & 13& 20 & 11\\
Qwen3-14B           & 13 & 15& 17 & 10& 12 & 13& 16 & 13\\
Qwen3-32B           & 11 & 13& 15 & 14& 12 & 15& 16 & 11\\
Qwen3-30B-A3B       & 11 & 15& 7 & 15& 10 & 15& 11 & 17\\
Qwen3-235B-A22B     & 12 & 14& 10 & 15& 11 & 14& 14 & 14\\
Llama-3.2-3B        & 15 & 10& 12 & 16& 12 & 16& 11 & 15\\
Llama-3.1-70B       & 14 & 13& 15 & 15& 13 & 15& 14 & 11\\
DeepSeek-V3.1       & 7 & 16& 9 & 15& 10 & 16& 12 & 17\\
Gemini-2.5-Flash    & 11 & 16& 8 & 15& 7 & 14& 5 & 16\\
Claude-4.5-Sonnet   & 11 & 14& 9 & 15& 9 & 15& 12 & 16\\
Kimi-K2             & 15 & 15& 12 & 13& 10 & 14& 11 & 12\\
GLM-4.5             & 12 & 14& 10 & 15& 10 & 15& 8 & 15\\
\midrule
\textbf{Total}& 191 & 208& 179 & 213& 185 & 206& 199 & 201\\
\cmidrule(lr){2-3} \cmidrule(lr){4-5} \cmidrule(lr){6-7} \cmidrule(lr){8-9}
& \multicolumn{2}{c}{\textbf{399}} & \multicolumn{2}{c}{\textbf{392}} & \multicolumn{2}{c}{\textbf{391}} & \multicolumn{2}{c}{\textbf{400}} \\
\bottomrule
\end{tabular}
\end{table*}

\begin{table*}[!h]
\centering
\caption{Breakdown statistics of the logical reasoning domain for LSR-Ben.}
\label{tab:logical_results}
\small 
\setlength{\tabcolsep}{1mm} 
\begin{tabular}{l c c c c c c c c c c}
\toprule
\multirow{2}{*}{\textbf{Model}} & \multicolumn{2}{c}{\textbf{CauseLogics}} & \multicolumn{2}{c}{\textbf{Analobench}} & \multicolumn{2}{c}{\textbf{LogiQA2}} & \multicolumn{2}{c}{\textbf{FOLIO}} & \multicolumn{2}{c}{\textbf{MIRAGE}} \\
\cmidrule{2-3} \cmidrule{4-5} \cmidrule{6-7} \cmidrule{8-9} \cmidrule{10-11}
& \textbf{Error} & \textbf{Correct} & \textbf{Error} & \textbf{Correct} & \textbf{Error} & \textbf{Correct} & \textbf{Error} & \textbf{Correct} & \textbf{Error} & \textbf{Correct} \\
\midrule
Gemma3-12B-IT              & 10 & 19& 17 & 12& 23 & 12& 21 & 13& 16 & 10\\
Gemma3-27B-IT              & 14 & 13& 16 & 14& 19 & 14& 18 & 12& 30 & 9\\
Qwen3-4B                   & 20 & 14& 19 & 10& 15 & 11& 12 & 15& 20 & 11\\
Qwen3-8B                   & 12 & 15& 13 & 12& 14 & 17& 10 & 14& 26 & 17\\
Qwen3-14B                  & 13 & 10& 12 & 10& 14 & 12& 12 & 16& 22 & 13\\
Qwen3-32B                  & 18 & 15& 18 & 12& 10 & 16& 12 & 12& 14 & 15\\
Qwen3-30B-A3B              & 19 & 15& 8 & 14& 2 & 10& 8 & 8& 0 & 4\\
Qwen3-235B-A22B            & 12 & 9& 6 & 17& 0 & 12& 9 & 15& 0 & 14\\
Llama-3.1-70B              & 15 & 13& 15 & 16& 24 & 12& 16 & 22& 36 & 8\\
Llama-3.2-3B               & 12 & 11& 19 & 17& 28 & 10& 31 & 9& 12 & 0\\
DeepSeek-v3.1              & 9 & 13& 17 & 8& 12 & 15& 15 & 10& 2 & 11\\
Gemini-2.5-Flash           & 16 & 13& 7 & 16& 2 & 15& 8 & 16& 1 & 14\\
Claude-4.5-Sonnet          & 0 & 18& 16 & 15& 16 & 15& 16 & 13& 14 & 26\\
Kimi-K2                    & 17 & 13& 12 & 14& 18 & 16& 13 & 12& 5 & 27\\
GLM-4.5                    & 13 & 11& 3 & 17& 2 & 13& 9 & 13& 4 & 19\\
\midrule
\textbf{Total}& 200 & 202 & 198 & 204 & 199 & 200 & 210 & 200 & 202 & 198\\
\cmidrule(lr){2-3} \cmidrule(lr){4-5} \cmidrule(lr){6-7} \cmidrule(lr){8-9} \cmidrule(lr){10-11}
 & \multicolumn{2}{c}{\textbf{402}} & \multicolumn{2}{c}{\textbf{402}} & \multicolumn{2}{c}{\textbf{399}} & \multicolumn{2}{c}{\textbf{410}} & \multicolumn{2}{c}{\textbf{400}} \\
\bottomrule
\end{tabular}
\end{table*}

\subsection{Limitations}
As far as we know, only two PRMs trained on broader reasoning domains (rather than only on the mathematical domain) are publicly available (VersaPRM and Qwen2.5-PRM-7B-PDDL), while the alternative OpenPRM remains closed-source. Furthermore, upon inquiry to the original authors, we were informed that the rental servers initially deployed for OpenPRM had been shut down, and the relevant data and model weights were not migrated in a timely manner; consequently, the OpenPRM model can no longer be retrieved. As a result, we only conducted experiments on VersaPRM and Qwen2.5-PRM-7B-PDDL among PRMs targeting board reasoning scenarios. If additional open-source PRMs for general reasoning domains are released in future work, we will extend our evaluation to include them accordingly. 

\subsection{Compute Resources}
In the process of benchmark construction and model evaluation, we employed four A100-SXM4-80GB GPUs.

\subsection{Detailed Evaluation Results}
\label{sec:detailevaluationresults}
\begin{table*}[!h]
\small
\centering
\caption{Evaluation results of three subdomains (Biology, Physics, and Chemistry) on LSR-Ben. We report accuracies on erroneous samples (samples with an erroneous reasoning step) and correct samples (samples without an erroneous reasoning step), together with their corresponding F1 scores.}
\label{tab:gr_ben_part1}
\setlength{\tabcolsep}{0.8mm}
\renewcommand{\arraystretch}{1.15}
\begin{tabular}{l ccc ccc ccc}
\toprule
\multirow{2}{*}{\textbf{Model}} 
& \multicolumn{3}{c}{\textbf{Biology}} 
& \multicolumn{3}{c}{\textbf{Physics}} 
& \multicolumn{3}{c}{\textbf{Chemistry}} \\
\cmidrule(lr){2-4} \cmidrule(lr){5-7} \cmidrule(lr){8-10}
& Erroneous & Correct & F1
& Erroneous & Correct & F1
& Erroneous & Correct & F1 \\
\midrule
\multicolumn{10}{c}{\textit{\textbf{Process Reward Models}}} \\
\midrule
ReasonEval-7B              & 7.9  & 84.6 & 14.4 & 15.1 & 65.7 & 24.5 & 18.4 & 53.4 & 27.3 \\
ReasonEval-34B             & 13.6 & 27.4 & 18.2 & 20.1 & 24.4 & 22.1 & 23.2 & 24.3 & 23.7 \\
Llemma-MetaMath-7B         & 8.9  & 28.8 & 13.6 & 10.6 & 68.5 & 18.4 & 14.6 & 51.9 & 22.8 \\
Llemma-PRM800K-7B          & 7.9  & 15.4 & 10.4 & 10.1 & 62.4 & 17.3 & 12.4 & 80.6 & 21.5 \\
Llemma-OPRM-7B             & 10.5 & 25.0 & 14.8 & 2.8  & 85.4 & 5.4  & 5.9  & 77.7 & 11.0 \\
Qwen2.5-Math-SCAN-Pro-7B   & 11.5 & 85.6 & 20.3 & 21.2 & 63.4 & 31.8 & 18.4 & 57.3 & 27.8 \\
Qwen2.5-Math-PRM800K-7B    & 9.4  & 94.7 & 17.1 & 12.8 & 83.1 & 22.3 & 16.8 & 74.3 & 27.3 \\
FoVer-Qwen2.5-7B           & 14.7 & 69.2 & 24.2 & 19.0 & 40.8 & 25.9 & 21.6 & 36.4 & 27.1 \\
Universal-PRM-7B           & 16.8 & 45.2 & 24.4 & 21.8 & 82.2 & 34.4 & 27.6 & 55.8 & 36.9 \\
Qwen2.5-Math-PRM-7B        & 13.1 & 90.4 & 22.9 & 21.8 & 83.6 & 34.6 & 20.5 & 77.7 & 32.5 \\
Qwen2.5-Math-PRM-72B       & 20.4 & 94.2 & 33.6 & 34.1 & 85.4 & 48.7 & 27.0 & 83.0 & 40.8 \\
Qwen2.5-PRM-7B-PDDL & 14.1 & 73.6 & 23.7 & 24.6 & 78.4 & 37.4 & 23.2 & 68.0 & 34.6 \\
VersaPRM (8B)              & 32.5 & 69.7 & 44.3 & 43.1 & 24.2 & 31.0 & 15.7 & 39.3 & 22.4 \\
\midrule
\multicolumn{10}{c}{\textit{\textbf{Open-source language models, prompted to identify reasoning errors}}} \\
\midrule
Llama-3.3-70B-Instruct     & 16.1 & 96.6 & 27.7 & 12.7 & 88.2 & 22.2 & 9.2 & 84.5 & 16.7  \\
Qwen3-4B                   & 23.4 & 70.0 & 35.1 & 16.6 & 70.6 & 26.9 & 13.0 & 70.5 & 22.0   \\
Qwen3-8B                   & 26.0 & 66.7 & 37.5 & 19.9 & 71.6 & 31.1 & 15.8 & 67.6 & 25.6   \\
Qwen3-14B                  & 24.0 & 93.2 & 38.1 & 25.4 & 82.9 & 38.9 & 17.9 & 76.3 & 29.0   \\
Qwen3-32B                  & 28.6 & 76.3 & 41.7 & 27.1 & 77.2 & 40.1 & 15.8 & 78.7 & 26.3   \\
Gemma-3-12B-it             & 31.8 & 42.5 & 36.4 & 22.1 & 39.3 & 28.3 & 21.2 & 37.7 & 27.1  \\
Gemma-3-27B-it             & 31.9 & 75.0 & 44.8 & 24.6 & 64.8 & 35.6 & 15.7 & 67.0 & 25.4   \\
Kimi-K2-Instruct-0905      & 39.3 & 59.6 & 47.4 & 38.5 & 78.4 & 51.7 & 30.3 & 73.3 & 42.9   \\
DeepSeek-v3.2              & 39.3 & 68.3 & 49.9 & 52.2 & 74.7 & 61.5 & 39.3 & 69.8 & 50.3  \\
\midrule
\multicolumn{10}{c}{\textit{\textbf{Proprietary language models}}} \\
\midrule
GPT-5.5-2026-04-24         & 37.2 & 64.9 & 47.3 & 41.3 & 72.8 & 52.7 & 30.8 & 75.2 & 43.7  \\
Gemini-3.8-flash           & 46.1 & 92.8 & 61.6 & 54.7 & 88.3 & 67.6 & 41.6 & 82.5 & 55.3  \\
\bottomrule
\end{tabular}
\end{table*}

\begin{table*}[!h]
\small
\centering
\caption{Evaluation results of three subdomains (Computer Science, Abductive reasoning, and Analogical reasoning) on LSR-Ben. We report accuracies on erroneous samples (samples with an erroneous reasoning step) and correct samples (samples without an erroneous reasoning step), together with their corresponding F1 scores.}
\label{tab:gr_ben_part2}
\setlength{\tabcolsep}{0.8mm}
\renewcommand{\arraystretch}{1.15}
\begin{tabular}{l ccc ccc ccc}
\toprule
\multirow{2}{*}{\textbf{Model}} 
& \multicolumn{3}{c}{\textbf{Computer Science}} 
& \multicolumn{3}{c}{\textbf{Abductive}} 
& \multicolumn{3}{c}{\textbf{Analogical}} \\
\cmidrule(lr){2-4} \cmidrule(lr){5-7} \cmidrule(lr){8-10}
& Erroneous & Correct & F1
& Erroneous & Correct & F1
& Erroneous & Correct & F1 \\
\midrule
\multicolumn{10}{c}{\textit{\textbf{Process Reward Models}}} \\
\midrule
ReasonEval-7B & 10.1 & 68.2 & 17.5 & 3.5 & 90.6 & 6.7 & 11.6 & 84.8 & 20.4 \\
ReasonEval-34B & 23.1 & 27.4 & 25.1 & 16.0 & 59.4 & 25.2 & 7.6 & 2.0 & 3.2 \\
Llemma-MetaMath-7B & 12.1 & 36.3 & 18.1 & 4.5 & 54.5 & 8.3 & 5.6 & 6.9 & 6.1 \\
Llemma-PRM800K-7B & 7.5 & 54.2 & 13.2 & 7.5 & 55.4 & 13.2 & 12.6 & 36.8 & 18.8 \\
Llemma-OPRM-7B & 7.0 & 43.3 & 12.1 & 9.5 & 16.8 & 12.1 & 12.1 & 5.9 & 7.9 \\
Qwen2.5-Math-SCAN-Pro-7B & 8.5 & 72.6 & 15.3 & 8.5 & 43.6 & 14.2 & 21.2 & 88.2 & 34.2 \\
Qwen2.5-Math-PRM800K-7B & 5.5 & 84.6 & 10.4 & 1.0 & 93.1 & 2.0 & 14.6 & 95.1 & 25.4 \\
FoVer-Qwen2.5-7B & 11.1 & 61.2 & 18.7 & 7.0 & 72.3 & 12.8 & 16.7 & 65.7 & 26.6 \\
Universal-PRM-7B & 19.1 & 58.2 & 28.8 & 10.0 & 40.1 & 16.0 & 22.7 & 71.6 & 34.5 \\
Qwen2.5-Math-PRM-7B & 11.6 & 78.1 & 20.1 & 2.0 & 82.2 & 3.9 & 26.3 & 90.7 & 40.7 \\
Qwen2.5-Math-PRM-72B & 30.7 & 94.0 & 46.2 & 4.5 & 83.2 & 8.5 & 33.8 & 91.2 & 49.4 \\
Qwen2.5-PRM-7B-PDDL & 17.7 & 64.2 & 27.6 & 4.0 & 69.3 & 7.6 & 30.3 & 82.8 & 44.4 \\
VersaPRM (8B) & 13.6 & 81.6 & 23.3 & 29.0 & 64.4 & 40.0 & 48.0 & 62.3 & 54.2 \\
\midrule
\multicolumn{10}{c}{\textit{\textbf{Open-source language models, prompted to identify reasoning errors}}} \\
\midrule
Llama-3.3-70B-Instruct & 17.1 & 93.5 & 28.9 & 4.5 & 92.6 & 8.6 & 15.6 & 98.0 & 26.9 \\
Qwen3-4B & 17.1 & 70.7 & 27.5 & 14.0 & 29.7 & 19.0 & 24.1 & 88.7 & 37.9 \\
Qwen3-8B & 24.6 & 65.2 & 35.7 & 11.5 & 74.3 & 19.9 & 41.2 & 70.9 & 52.1 \\
Qwen3-14B & 18.1 & 77.6 & 29.3 & 14.5 & 82.7 & 24.7 & 35.7 & 83.2 & 50.0 \\
Qwen3-32B & 23.1 & 71.6 & 35.0 & 13.0 & 72.3 & 22.0 & 41.2 & 76.3 & 53.5 \\
Gemma-3-12B-it & 23.1 & 39.3 & 29.1 & 13.0 & 8.4 & 10.2 & 28.1 & 16.3 & 20.6 \\
Gemma-3-27B-it & 21.1 & 60.2 & 31.2 & 34.1 & 32.2 & 33.1 & 37.2 & 72.4 & 49.1 \\
Kimi-K2-Instruct-0905 & 42.7 & 64.2 & 51.3 & 35.5 & 84.2 & 49.9 & 50.0 & 42.2 & 45.7 \\
DeepSeek-v3.2 & 49.0 & 72.6 & 58.5 & 61.5 & 85.2 & 71.4 & 30.8 & 67.2 & 42.2 \\
\midrule
\multicolumn{10}{c}{\textit{\textbf{Proprietary language models, prompted to identify reasoning errors}}} \\
\midrule
GPT-5.5-2026-04-24 & 49.2 & 67.7 & 57.0 & 40.0 & 71.3 & 54.4 & 26.3 & 73.0 & 38.6 \\
Gemini-3.8-flash & 57.3 & 88.1 & 69.4 & 58.0 & 92.6 & 71.3 & 38.9 & 95.1 & 55.2 \\
\bottomrule
\end{tabular}
\end{table*}

\begin{table*}[!h]
\small
\centering
\caption{Evaluation results of three subdomains (Mixed reasoning, Deductive reasoning, and Inductive reasoning) on LSR-Ben. We report accuracies on erroneous samples (samples with an erroneous reasoning step) and correct samples (samples without an erroneous reasoning step), together with their corresponding F1 scores.}
\label{tab:gr_ben_part3}
\setlength{\tabcolsep}{0.8mm}
\renewcommand{\arraystretch}{1.15}
\begin{tabular}{l ccc ccc ccc}
\toprule
\multirow{2}{*}{\textbf{Model}} 
& \multicolumn{3}{c}{\textbf{Mix}} 
& \multicolumn{3}{c}{\textbf{Deductive}} 
& \multicolumn{3}{c}{\textbf{Inductive}} \\
\cmidrule(lr){2-4} \cmidrule(lr){5-7} \cmidrule(lr){8-10}
& Erroneous & Correct & F1
& Erroneous & Correct & F1
& Erroneous & Correct & F1 \\
\midrule
\multicolumn{10}{c}{\textit{\textbf{Process Reward Models}}} \\
\midrule
ReasonEval-7B              & 7.0  & 71.0 & 12.7 & 6.2  & 66.5 & 11.3 & 24.6 & 20.5 & 22.4 \\
ReasonEval-34B             & 40.5 & 4.5  & 8.1  & 13.7 & 44.5 & 21.0 & 38.9 & 1.5  & 2.9  \\
Llemma-MetaMath-7B         & 1.0  & 95.0 & 2.0  & 6.6  & 39.5 & 11.4 & 13.3 & 28.5 & 18.1 \\
Llemma-PRM800K-7B          & 13.0 & 44.5 & 20.1 & 10.4 & 17.5 & 13.1 & 13.3 & 24.0 & 17.1 \\
Llemma-OPRM-7B             & 3.0  & 33.5 & 5.5  & 8.5  & 26.5 & 12.9 & 27.1 & 1.0  & 1.9  \\
Qwen2.5-Math-SCAN-Pro-7B   & 4.5  & 90.5 & 8.6  & 7.1  & 85.0 & 13.1 & 30.5 & 8.0  & 12.7 \\
Qwen2.5-Math-PRM800K-7B    & 9.5  & 91.5 & 17.2 & 5.2  & 95.5 & 9.9  & 32.0 & 43.5 & 36.9 \\
FoVer-Qwen2.5-7B           & 14.1 & 86.5 & 24.2 & 13.8 & 63.5 & 22.7 & 32.2 & 27.3 & 29.5 \\
Universal-PRM-7B           & 38.0 & 50.0 & 43.2 & 19.0 & 26.5 & 22.1 & 44.3 & 11.0 & 17.6 \\
Qwen2.5-Math-PRM-7B        & 16.5 & 86.0 & 27.7 & 12.8 & 83.0 & 22.2 & 38.9 & 30.0 & 33.9 \\
Qwen2.5-Math-PRM-72B       & 23.0 & 90.5 & 36.7 & 23.2 & 87.5 & 36.7 & 43.3 & 30.5 & 35.8 \\
Qwen2.5-PRM-7B-PDDL & 22.1 & 75.5 & 34.2 & 14.3 & 73.5 & 23.9 & 40.6 & 18.7 & 25.6 \\
VersaPRM (8B)              & 36.0 & 47.5 & 41.0 & 12.3 & 10.5 & 11.3 & 33.0 & 48.5 & 39.3 \\
\midrule
\multicolumn{10}{c}{\textit{\textbf{Open-source language models, prompted to identify reasoning errors}}} \\
\midrule
Llama-3.3-70B-Instruct      & 10.5 & 93.5 & 18.9 & 12.8 & 79.5 & 22.0 & 17.7 & 76.5 & 28.8 \\
Qwen3-4B                    & 29.0 & 61.5 & 39.4 & 14.7 & 71.5 & 24.4 & 21.7 & 55.0 & 31.1 \\
Qwen3-8B                    & 33.5 & 56.0 & 41.9 & 19.4 & 65.0 & 29.9 & 25.1 & 70.5 & 37.0 \\
Qwen3-14B                   & 26.0 & 83.5 & 39.6 & 17.1 & 81.5 & 28.2 & 20.2 & 80.0 & 32.2 \\
Qwen3-32B                   & 35.0 & 65.5 & 45.6 & 22.3 & 71.5 & 34.0 & 20.2 & 78.0 & 32.1 \\
Gemma-3-12B-it              & 34.0 & 28.0 & 30.7 & 14.2 & 45.5 & 21.7 & 20.2 & 27.0 & 23.1 \\
Gemma-3-27B-it              & 26.0 & 60.5 & 36.4 & 12.8 & 69.0 & 21.6 & 27.1 & 49.0 & 34.9 \\
Kimi-K2-Instruct-0905       & 52.5 & 44.0 & 47.9 & 27.0 & 67.0 & 38.5 & 60.1 & 74.5 & 66.5 \\
DeepSeek-v3.2               & 44.0 & 64.5 & 52.3 & 34.1 & 65.5 & 44.9 & 63.5 & 72.0 & 67.5 \\
\midrule
\multicolumn{10}{c}{\textit{\textbf{Proprietary language models, prompted to identify reasoning errors}}} \\
\midrule
GPT-5.5-2026-04-24          & 54.3 & 68.5 & 60.6 & 25.7 & 56.0 & 35.2 & 52.5 & 73.2 & 61.1 \\
Gemini-3.8-flash            & 56.8 & 88.0 & 69.0 & 37.3 & 68.5 & 48.3 & 70.3 & 82.8 & 76.0 \\
\bottomrule
\end{tabular}
\end{table*}

\end{document}

%% file: math_commands.tex
\usepackage{amsmath,amsfonts,bm}

\def\eqref#1{equation~\ref{#1}}

\def\1{\bm{1}}

\DeclareMathAlphabet{\mathsfit}{\encodingdefault}{\sfdefault}{m}{sl}
\SetMathAlphabet{\mathsfit}{bold}{\encodingdefault}{\sfdefault}{bx}{n}

